\documentclass[10pt,twocolumn,letterpaper]{article}

\usepackage{iccv}      %

\definecolor{iccvblue}{rgb}{0.21,0.49,0.74}
\usepackage[pagebackref,breaklinks,colorlinks,allcolors=iccvblue]{hyperref}
\usepackage{comment}
\def\paperID{14225} %
\def\confName{ICCV}
\def\confYear{2025}

\newcommand{\xhdr}[1]{\vspace{0em}\noindent{{\bf #1.}}}
\newcommand{\method}{\textsc{DoPTA}}
\usepackage{float}
\usepackage{colortbl} %
\usepackage{xcolor}   %

\title{\method: Improving Document Layout Analysis using Patch-Text Alignment}

\author{
Nikitha SR\thanks{Equal contribution. Correspondence to \texttt{nikithasr@adobe.com}.} \hspace{15pt}
Tarun Ram Menta\footnotemark[1] \hspace{15pt}
Mausoom Sarkar\\
Media and Data Science Research Lab, Adobe \\
{\tt\small \{nikithasr, tarunramm, msarkar\}@adobe.com}
}

\begin{document}
\maketitle

\begin{abstract}

The advent of multimodal learning has brought a significant improvement in document AI. Documents are now treated as multimodal entities, incorporating both textual and visual information for downstream analysis. However, works in this space are often focused on the textual aspect, using the visual space as auxiliary information. While some works have explored pure vision based techniques for document image understanding, they require OCR identified text as input during inference, or do not align with text in their learning procedure. Therefore, we present a novel image-text alignment technique specially designed for leveraging the textual information in document images to improve performance on \textit{visual} tasks. Our document encoder model~\method~- trained with this technique demonstrates strong performance on a wide range of document image understanding tasks, without requiring OCR during inference. Combined with an auxiliary reconstruction objective,~\method~consistently outperforms larger models, while using significantly lesser pre-training compute.~\method~also sets new state-of-the art results on D\textsuperscript{4}LA, and FUNSD, two challenging document visual analysis benchmarks


\end{abstract}

\section{Introduction}
\label{sec:intro}
Document images are a rich source of information in the modern age. Compared to natural images, document images often have a complex structure composed of high-frequency details like text, tables, figures, charts, etc. In addition, a document usually includes rich textual information and can be of various types (scientific paper, form, resume, etc.), each with its unique combinations of elements and layouts. This makes Visual Document Understanding (VDU) an important, and challenging task. VDU encompasses a wide variety of tasks, including but not limited to  classification~\cite{harley2015rvl}, layout analysis~\cite{zhong2019publaynet, pfitzmann2022doclaynet, li2020docbank, da2023vgt, huang2019icdar_table}, information extraction~\cite{park2019cord, stanislawek2021kleister}, and question answering~\cite{mathew2021docvqa, mathew2022infographicvqa}. Each of these tasks requires inspection of the document image at multiple levels of granularity. Additionally, the rich semantic structure of a document cannot be modeled by text or vision alone. The layout of text and different objects in the document, the appearance of text in different sections (font, color, size), and visual elements such as figures, tables, etc. make \textit{holistic} document understanding a complex and involved task. As such, this necessitates the careful design of special architectures and objectives for effective learning, departing from the general natural image representation learning methods. 
\begin{figure}
    \centering
    \scalebox{0.9}{\includegraphics[width=7cm]{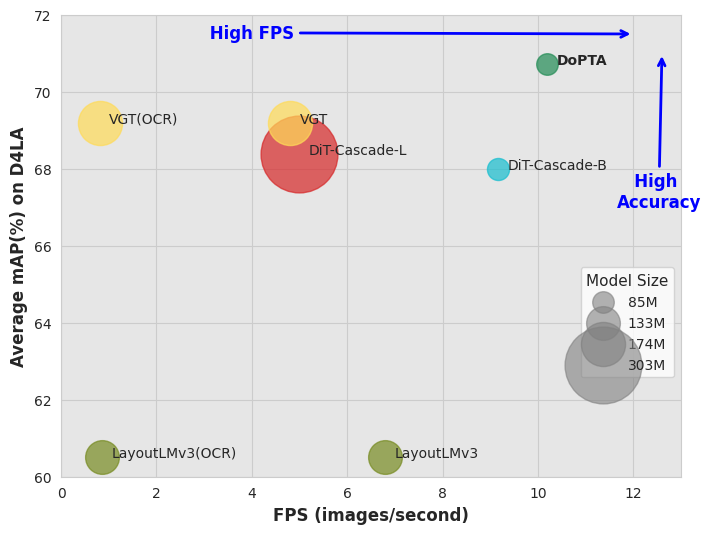}}
    \caption{Our method achieves superior FPS due to the OCR free inference setting while also setting SOTA mAP as compared to several existing methods. Model(OCR) denotes the FPS when OCR parsing is taken into account for computing inference time.}
    \label{fig:fps}
\end{figure}
The aforementioned reasons highlight the necessity of multimodal modeling for effective document understanding. Currently, transformer architecture has evolved as a ubiquitous framework capable of modeling multiple modalities and has naturally been applied to VDU as well. Most works~\cite{layoutlm, xu2020layoutlmv2, huang2022layoutlmv3, appalaraju2021docformer, tang2023udop} use a unified transformer approach, wherein image, text, and layout information are processed by a single multi-modal transformer, which is pre-trained with a variety of objectives. The unified transformer based methods focus more on textual information, and treat visual information as secondary. They require extraction of text from a document image using standard OCR techniques, which is later modeled by the unified transformer for downstream tasks. This 2-stage paradigm has two key issues - inflexibility and latency of the OCR pipelines, and error propagation from OCR extraction to downstream tasks. Methods such as Donut~\cite{donut} instead model both OCR extraction and subsequent understanding of the document in a single end-to-end approach using an encoder-decoder transformer model. 

Although these approaches achieve strong performance on semantic tasks such as document question answering~\cite{mathew2021docvqa} and information extraction~\cite{stanislawek2021kleister}, they fall behind on visual tasks such as document layout analysis, as their primary focus is on modeling the textual features. In this field, state-of-the-art results have been achieved by DiT~\cite{li2022dit} and VGT~\cite{da2023vgt}. DiT approaches document image understanding in a self-supervised fashion, wherein masked image patches are reconstructed through a ViT encoder, and matched to the tokens from a pre-trained dVAE. VGT builds upon this, adding a Grid Transformer (GiT) to infuse layout information into the image representations. While these methods achieve strong results, we argue that the semantic information from text in the image can be a strong factor in improving the layout understanding. Inspired by the power of contrastive language-image training in representation learning~\cite{radford2021clip} and fine-grained image-text alignment techniques like FILIP~\cite{yao2022filip}, we specially design a patch-text alignment loss for documents, using IoU to guide the model to learn effective representations that are semantically and structurally rich. Our key contributions can be summarized as follows:
\begin{itemize}
    \item We introduce a \textbf{novel patch-text alignment objective} guided by the IoU between text bounding boxes and image regions specially designed for document images, which effectively leverages the textual information in images to improve VDU. This objective bridges the gap between existing text-centric and vision-centric objectives, effectively leveraging both textual and visual data.
    \item We further build upon this, and propose\textbf{~\method}, a strong document image encoder trained on our objective in conjunction with existing self-supervised learning objectives for images.
    \item We rigorously evaluate~\method~on a variety of document understanding tasks to prove the efficacy of the learned representations in downstream tasks.~\method~is able to achieve strong performance across our evaluation benchmarks, while requiring less pre-training steps than the existing state-of-the-art. 
\end{itemize}

\section{Related Work}
\label{sec:related_work}
\begin{figure*}
    \centering
    \includegraphics[width=0.95\linewidth]{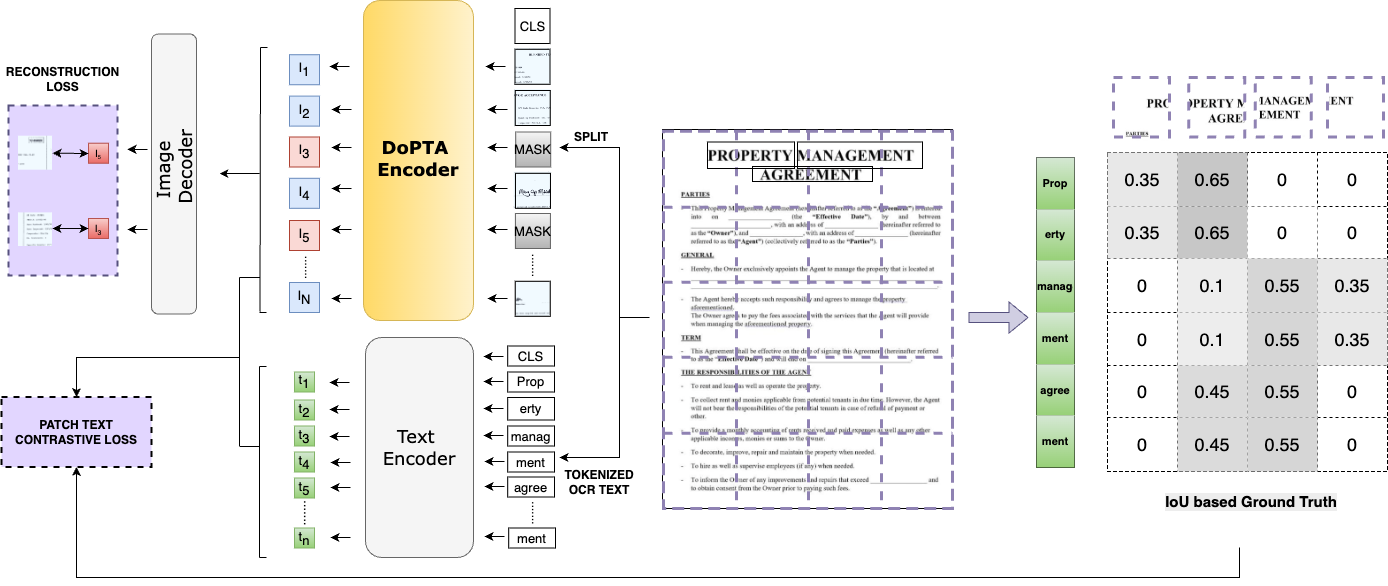}
    \caption{Pre-training of~\method. Only the image encoder is required for downstream usage. Refer section Sec.~\ref{sec:method} for details.}
    \label{fig:main-method}
\end{figure*}

\xhdr{Self-Supervised Image Representation Learning}
Learning effective visual representations without human supervision is crucial for leveraging the large amounts of unlabelled image data available on the web, and has emerged as a powerful pre-training paradigm for strong vision backbones without the need for large-scale labeled datasets like ImageNet~\cite{russakovsky2015imagenet}. MoCO~\cite{he2020moco}, SimCLR~\cite{chen2020simclr, chen2020simclrv2}, and their variants propose contrastive learning for learning effective representations by reducing the distance between representations of different augmented views of the same image and increasing the distance between representations of augmented views from different images. BYOL~\cite{grill2020byol} removes the need for large in-batch negatives and image augmentations by bootstrapping the outputs of a network to serve as targets for an enhanced representation. The emergence of Vision Transformers~\cite{dosovitskiy2021vit} which split the image into small patches to input to a bi-directional transformer encoder has inspired a slew of new learning methods. MAE~\cite{he2022mae} and BEIT~\cite{bao2021beit} learn visual representations by reconstructing masked image patches. Finally, methods such as DINO~\cite{caron2021dino} and DINOv2~\cite{oquab2023dinov2} align image crops with their global representations, using a distillation approach to achieve more fine-grained image representations. 
\\\\
\xhdr{Vision-Language Pre-training}
Works such as CLIP~\cite{radford2021clip}, ALIGN~\cite{jia2021align}, and more recently SigLIP~\cite{zhai2023sigmoidlosslanguageimage} show the effectiveness of using language to learn visual representations, with the help of large scale image-text datasets such as YFCC100M~\cite{thomee2016yfcc100m}, JFT-300M~\cite{sun2017jft300m}, CC12M~\cite{changpinyo2021cc12m}, LAION~\cite{schuhmann2022laion}. The core technique of these models lies in the global contrastive alignment of the images and texts through a dual-stream model, with a vision encoder, and a text encoder. While these approaches enable strong zero-shot and few-shot performance, they lack fine-grained representations, because of the global alignment objective that they use. Fine-grained representations through vision-language alignment has been explored in FILIP~\cite{yao2022filip}, SPARC~\cite{bica2024improvingfinegrainedunderstandingimagetext}, GLIP~\cite{li2022groundedlanguageimagepretraining}, UNITER~\cite{chen2020uniter} and VL-BERT~\cite{Su2020VL-BERT:}. These works use deep cross-modal fusion to align text to local image regions, show improved performance on tasks like object detection and are the closest to our proposed approach. Global image-text alignment is not well applicable to the document image setting, as short language captions fail to capture the complexity and details of dense, text-rich documents. Additionally, fine-grained representations are of utmost significance in document understanding tasks, where most details are small and cannot be detected by global alignment, which is what we explore in this work. 
\\\\
\xhdr{Document Image Understanding}
Visual document understanding(VDU) requires careful design of the objectives, owing to the unique structure of these images. The majority of approaches in this field can be categorized into two sub-categories based on the use/non-use of OCR as an input. i) \textit{OCR-Based methods} include BiVLDoc~\cite{luo2022bivldoc}, LayoutLM~\cite{layoutlm, xu2020layoutlmv2, huang2022layoutlmv3}, DocFormer~\cite{appalaraju2021docformer}, BROS~\cite{hong2022bros}, VL-BERT~\cite{Su2020VL-BERT:}, UDOP~\cite{tang2023udop}, VGT~\cite{da2023vgt}, TILT~\cite{powalski2021tilt}, M2Doc\cite{zhang2024m2doc} and UDOC. These works utilize off-the-shelf OCR methods to parse the text and bounding boxes from a document image. The textual and image features are later combined through early or late fusion, using a joint transformer encoder to produce the final representations. Different variants of objectives such as masked image modeling (MIM), masked language modeling (MLM), and image-language alignment are proposed in these papers. However, these works require OCR as an input during inference. ii) \textit{OCR-Free methods} such as Donut~\cite{donut}, DiT~\cite{li2022dit}, and StructTextv2~\cite{yu2023structextv2} instead aim to learn visual features in the absence of OCR as an input during inference, though it may be utilized as a target during pre-training. Donut uses a transformer encoder-decoder architecture with OCR parsing as its pre-training task. On the other hand, DiT learns image features in the absence of any OCR ground truth, by aligning image patches with learned tokens from a discrete VAE tokenizer. StructTextv2 uses a dual objective of image reconstruction and text prediction of masked-out regions. Our work falls into the second category.

\section{Methodology}
\label{sec:method}

We now present~\method, a novel pre-training method for learning document image representations with strong semantic and structural understanding. The key component of~\method~is the introduction of a novel fine-grained image-text contrastive alignment objective for document images. This loss imbibes textual-semantic information into the image representations, leading to better structural understanding through the semantics. Despite the strong performance demonstrated by this loss, as shown in Sec.~\ref{sec:ablations},~\method~also includes an image reconstruction loss to incorporate additional structural information. We present a detailed description of our losses and architecture in the following section. 

\subsection{Model architecture}

\label{sec:model-architecture}
\begin{figure}[t]
    \centering
    \scalebox{0.75}{    \includegraphics[width=\linewidth]{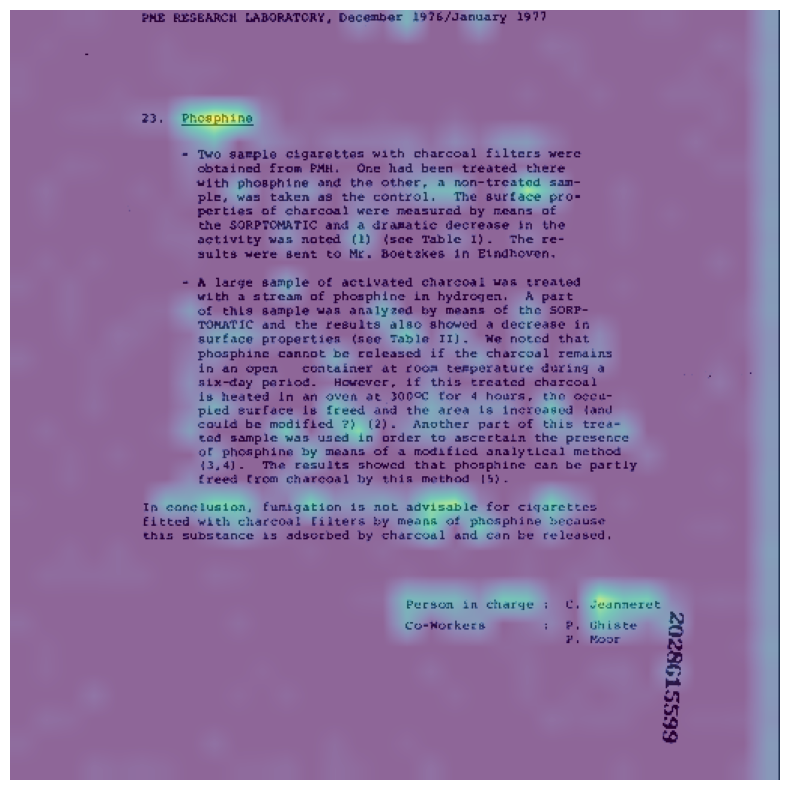}}
    \caption{Heatmap visualisation of the normalised dot product similarity of image region embeddings with the text embedding for the token `phosphine' taken from~\method model. Additional qualitative results are presented in Appendix~\ref{app:examples}.}
    \label{fig:attention_map}
\end{figure}

The pre-training stage of~\method~consists of three components - i) ~\method~Encoder, ii) Text Encoder, iii) Image Decoder. The latter two components are only required during pre-training, and only the~\method~encoder is used for downstream evaluations. Figure~\ref{fig:main-method} shows the main architectural components of our pre-training. We present qualitative examples of the effect of our patch-text alignment loss in Fig.~\ref{fig:attention_map}. 
\\\\
\xhdr{Image Encoder}
The \textit{~\method~encoder} (E$_I$) follows Vision Transformer~\cite{dosovitskiy2021vit}. We reshape an image $\mathbf{x} \in \mathbb{R}^{H \times W \times C}$ into a sequence of flattened 2D patches $\mathbf{x}_p \in \mathbb{R}^{N \times (P^2 \cdot C)}$, where $(H, W)$ is the resolution of the original image, $C$ is the number of channels, $(P, P)$ is the resolution of each image patch, and $N= HW/P^2$ is the resulting number of patches. We randomly mask out a fraction $M$ of the patches by replacing them with a learned [MASK] embedding, which is later reconstructed by the \textit{image decoder}. Learnable positional embeddings are added to each patch before passing it as input to the transformer. The per-patch embeddings are aligned with both textual and visual information to ensure a rich representation. 
\\\\
\xhdr{Text Encoder}
The text encoder (E$_T$) is a transformer model. We extract the text with corresponding bounding boxes from each document image using an off-the-shelf OCR engine. The entire set of texts is concatenated in reading order and tokenized as a single string. We truncate the tokenized string at a maximum sequence length of $L_T$ tokens and add learnable positional embeddings before passing through the text encoder model. We use the per-token embeddings at the output layer for aligning vision features. 
\\\\
\xhdr{Image Decoder}
Following MAE~\cite{he2022mae}, we model the image decoder (D$_I$) as a shallow 2 layer transformer model which maps the latent representations back to pixels. The last layer of the decoder is a linear projection with the number of output channels being equal to the number of pixel values in a patch. The input to the decoder is the full set of patches encoded by the image encoder (including both masked and unmasked patches). The decoder learns to reconstruct the pixels of the masked regions using the embeddings of the surrounding patches as context.

\subsection{Fine-Grained Image-Text Alignment}
\label{sec:our-loss}
Contrastive image-text learning~\cite{radford2021clip, jia2021align} is a powerful paradigm for learning cross-modal representations that can be decoupled for downstream uses. Models following this paradigm train unimodal dual encoders with images and a global description of the image in the form of text captions. Though this can be naturally extended to text-rich documents, modeling large-scale document images with global contrastive learning is sub-optimal. The positional layout of text in documents is of great importance. Hence, we propose a novel patch-text alignment objective for document image pre-training. We extend fine-grained contrastive approaches like~\cite{yao2022filip, li2022glip} and specially design our loss to suit the document domain. In particular, our patch-text alignment technique leverages the \textit{exact position} of text present in documents, using an IoU guided loss to achieve a high degree of understanding.

 The~\method~encoder (E$_I$) produces a set of patch level embeddings $\{X^I_i\}^{N}_{i=1}$  for the $N = HW/P^2$ patches of the image. The tokenized OCR text of the image is encoded in the reading order through the text encoder (E$_T$) to generate a set of $\{X^T_i\}^D_{i=1}$ text encodings where D is the predefined context length of the text encoder. We define a per image \textit{TextToPatch matching loss}  which is an asymmetric cross-entropy loss between each text token and the set of all image patches. The \textit{TextToPatch} contrastive loss $\mathcal{L}_i$ for a text token $X^T_i$ is given by,
 \begin{equation}
     \mathcal{L}_i(X^T_i, \{X^I_j\}^N_{j=1}) = -\sum_{j=1}^N \textit{Y}(T_i,I_j)\log{\frac{exp(\lambda \cdot s_{i,j})}{\sum_{k=1}^N exp(\lambda \cdot s_{i,k})}}
 \end{equation}
 where $\lambda$ is a learnt scaling factor and $s_{i,j} := X_i^T\cdot X_j^I$ is the dot product similarity between the $i^{th}$ text embedding and $j^{th}$ image patch embedding. The ground truth probability $\textit{Y}(T_i,I_j)$
 for a text token $T_i$ and an image patch $I_j$ is: 
\begin{equation}
    \textit{Y}(T_i,I_j) = \frac{|bbox(I_j) \cap bbox(T_i)|}{|bbox(T_i)|}
\end{equation}
where $bbox(.)$ is the bounding box of the enclosed entity. In simple terms, we enforce the \emph{probability distribution of the similarity between text embedding and the image embeddings to match (directly correlate to) the distribution of text area across image patches}. A pictorial representation of the ground truth generation is also shown in Figure~\ref{fig:main-method}. The overall \textit{TextToPatch} contrastive loss $\mathcal{L}_{TP}$ is obtained by averaging across the text token losses.
\begin{equation}
    \mathcal{L}_{TP} = \frac{1}{D}\sum\mathcal{L}_i
\end{equation}
This smoothened contrastive loss infuses strong textual and text-structure information into the visual representations. 

\subsection{Image Reconstruction Loss}
\label{sec:reconstruction-loss}
While the \textit{TextToPatch} contrastive loss takes care of the textual portions of the image, documents also constitute other visual components like graphs or diagrams which do not contain text. To learn their representations in a better fashion an image reconstruction loss following MAE~\cite{he2022mae} is also included. A certain fraction $M$ of the image patches are replaced with a learned [MASK] token while being passed through the ~\method~ encoder. Since a large fraction of the images are white space, these patches are never masked, so as to make the reconstructions non-trivial. The patch embeddings $\{X^I_i\}^{N}_{i=1}$ obtained from the~\method~encoder are combined with learnable positional embeddings and passed through the image decoder. Each output embedding $\{D^I_i\}^{N}_{i=1}$ is a vector of the linearised pixel values of the patch. The masked patch embeddings are reshaped to create a reconstructed image patch. The \textit{Reconstruction} loss $\mathcal{L}_{R}$ calculates the mean squared error (MSE) between the reconstructed patch and the original patch in normalized pixel space. The combined loss computed for each image is then given by,
\begin{equation}
    \mathcal{L} = \mathcal{L}_{TP} + \lambda\mathcal{L}_{R}
\end{equation}
where $\lambda$ assumes values from $\{0,1\}$ depending on the usage of the \textit{reconstruction} loss.

    

\section{Experiments}
\label{sec:experiments}
 Next, we present experimental results to show the effectiveness of image features produced by~\method~on a variety of document tasks, including document image classification (Sec.~\ref{sec:document-image-classification}), document layout analysis (Sec.~\ref{sec:document-layout-analysis}), and text detection (Sec.~\ref{sec:text-detection}). Evaluations show that our model \emph{achieve state-of-the-art results} in multiple tasks, \emph{outperforming larger models}, while adopting a significantly \emph{shorter pre-training schedule}.

\begin{figure}
    \centering
    \scalebox{0.8}{\includegraphics[width=\linewidth]{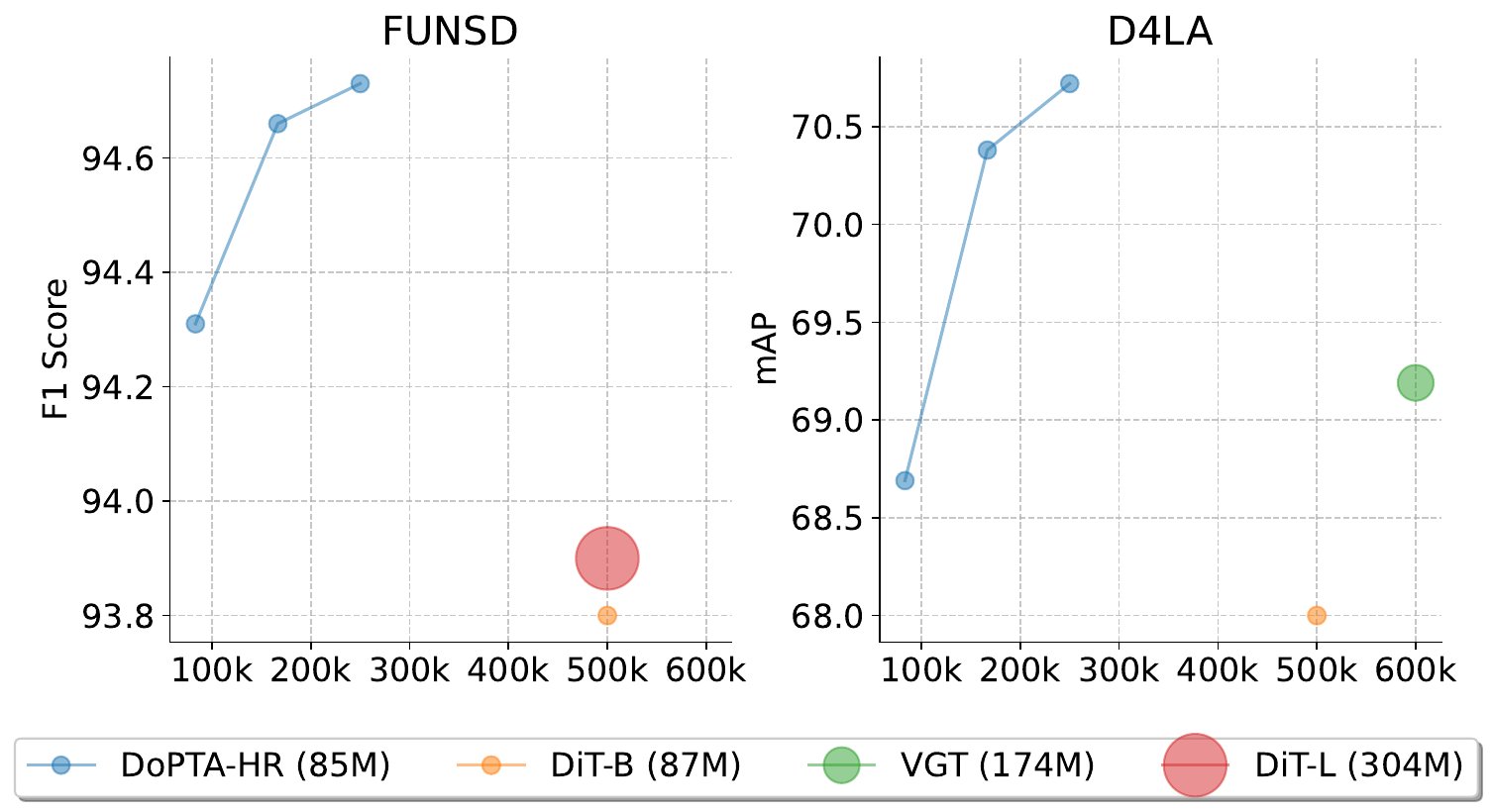}}
    \caption{Results of~\method~and existing SOTA document encoder models.~\method~outperforms other methods on multiple benchmarks, despite having less parameters, and a significantly shorter pre-training schedule. Refer to Sec.~\ref{sec:experiments} for more details of individual benchmarks}
    \label{fig:main_results}
\end{figure}

\subsection{Implementation and Pre-Training}
We pretrain~\method~on the IIT-CDIP~\cite{cdip2022} dataset. This dataset contains 42M pages of black-and-white document images containing rich text. We extract word-level OCR text and their bounding boxes using EasyOCR~\cite{JadedAI} pipeline and use random cropping as the image augmentation. Though we carefully ensure the quality of the extracted OCR through filtering, some errors in extracted OCR do persist. In Appendix~\ref{app:pixparse}, we explore the use of a PDF dataset which circumvents this issue. However, we choose the CDIP dataset for pre-training due to its larger scale, and to maintain parity with the baselines, which pre-train on the same dataset. An important point to note is that our method \textbf{does not} require any OCR input during inference. 

We use a mix of padded (aspect ratio preserving) and square-cropped images during training to ensure good downstream performance in all settings. We follow the architectural choices of 'CLIP-ViT-B/16', with our DOPTA encoder and text encoder being 12-layer transformer with 8 attention heads. The hidden (intermediate) sizes are 768 (3072) for the DOPTA encoder and 512 (2048) for the text encoder. Both models are initialized using the 'CLIP-ViT-B/16' weights. The context length of the text encoder is set to $512$ by linear interpolating the learnt CLIP positional embeddings. As discussed in Sec.~\ref{sec:model-architecture}, we adopt a lightweight 2-layer transformer image decoder each with $8$ attention heads. We train~\method~with image resolution $512\times512$. The model uses a patch size of 16, a global batch size of $2048$, dropout of $0.1$, and learning rate of $1e-3$. The masking ratio $M$ for reconstruction is set to $0.6$. We train~\method~ for 15 epochs ($\approx$250k steps). This is a significantly shorter pre-training schedule compared to other works like DiT~\cite{li2022dit}, LayoutLMv3~\cite{huang2022layoutlmv3}, and VGT~\cite{da2023vgt}, which are pretrained for 500k steps or more. 

\subsection{Document Image Classification}
\label{sec:document-image-classification}
We use the RVL-CDIP benchmark to evaluate the classification performance of~\method. The benchmark consists of 400K document images split into 320K train, 40K validation and 40K test images. It consists of $16$ different classes like advertisement, email, form, scientific publication, etc. To perform classification, we obtain a single representation embedding per image by average pooling the patch-wise embeddings and directly applying a linear classification head on top. We evaluate~\method~encoder by finetuning for 100 epochs on the training set as done in DiT~\cite{li2022dit}. We use AdamW optimiser with a learning rate of $1e-3$, a global batch size of 1024 and perform gradient clipping with a value of $0.1$. 

\xhdr{Baselines and Results} We compare and report results of classification accuracy on the test set in Table~\ref{tab:rvl-cdip}. We consider two categories of methods - i) \textit{OCR-based methods} which rely on the OCR-identified text in the image as input, and ii) \textit{OCR-Free methods} which treat document image classification purely in the image domain. All results are taken from the DiT\cite{li2022dit}, except Donut which we finetune ourselves. Donut~\cite{donut} utilizes an encoder-decoder architecture for end-to-end OCR extraction. To test the performance of Donut in the image domain, we utilize the image encoder alone and evaluate it using the aforementioned setup. While the OCR-based methods achieve the highest performance in this category, we find that ~\method~ outperform all OCR-free methods. It is notable that~\method~ outperforms even the DiT-L model, despite having $<1/3^{rd}$ the parameters, and a much shorter pre-training schedule (250k steps in our case as compared to 500k steps for DiT-L).

\begin{table}
  \centering
  \setlength{\tabcolsep}{2pt}
  \scalebox{0.8}{
  \begin{tabular}{lccc}
    \toprule
    Model & Resolution & Accuracy & \#Param\\
    \midrule
    \textit{Text-Based Methods} & & & \\
    BERT & - & 89.81 & 110M \\
    LayoutLMv3~\cite{huang2022layoutlmv3} & - & 95.44 &  133M\\
    DocFormer & - & \textbf{96.17} & 183M \\
    \midrule
    \textit{Image Encoders} & & & \\
    EAML~\cite{bakkali2023eamlensembleselfattentionbasedmutual} & 229 & 90.81 & \\
    DeiT-B~\cite{touvron2021trainingdataefficientimagetransformers} & 224 & 90.32 & 87M \\
    BEiT-B ~\cite{bao2021beit} & 224 & 91.09 & 87M\\
    MAE-B~\cite{he2022mae} & 224 & 91.42 & 87M\\
    NasNet$_{Large}$~\cite{bakkali2020visual} & 224  & 91.45 & 88M \\
    DiT-B~\cite{li2022dit} & 224 & 92.11 & 87M\\
    DiT-L~\cite{li2022dit} & 224 & 92.69 & 304M \\
    Donut-Encoder~\cite{donut} & 512 & 93.37 & 71M \\
    StructTexTv2-Small~\cite{yu2023structextv2}& 960 & 93.4 & 28M \\ 
    
     \midrule
     \method & 512& \textbf{94.12} & 85M \\ 
    \bottomrule
  \end{tabular}}
  \caption{
  Classification Accuracy on RVL-CDIP Test set. Higher is better. Best result in each category is indicated in \textbf{bold}}
  \label{tab:rvl-cdip}
\end{table}

\begin{table}
    \centering
      \setlength{\tabcolsep}{5pt}
      \scalebox{0.68}{
      \begin{tabular}{lc|ccccc|c}
      \toprule
      Model & Parameters & Text & Title & List & Table & Figure & Overall \\
      \midrule
      ResNeXt~\cite{xie2017aggregatedresidualtransformationsdeep} & - & 91.6 & 84.5 & 91.8 & 97.1 & 95.2 & 92.0 \\
      DeiT-B~\cite{touvron2021trainingdataefficientimagetransformers} & - & 93.4 & 87.4 & 92.1 & 97.2 & 95.7 & 93.2 \\
      BEiT-B~\cite{bao2021beit} & - & 93.4 & 86.6 & 92.4 & 97.3 & 95.7 & 93.1 \\
      MAE-B~\cite{he2022mae} & - & 93.3 & 86.5 & 91.8 & 97.3 & 95.9 & 93.0 \\
      UDoc~\cite{} & - & 93.9&88.5&93.7&97.3&96.4&93.9\\
      Donut-Encoder~\cite{donut} & 72M & 93.9 & 87.5 & 95.2 & 97.6 &  96.9 & 94.2 \\
      M2Doc$^*$~\cite{zhang2024m2doc} &-& 94.3&88.7&95.2&97.3&96.7&94.5\\
      DiT-B~\cite{li2022dit} & 87M & 94.4 & 88.9 & 94.8 & 97.6 & 96.9 & 94.5 \\
      DiT-L~\cite{li2022dit} & 304M & 94.4 & 89.3 & 96.0 & 97.8 & 97.2 & 94.9 \\
      VGT$^*$~\cite{da2023vgt} & 174M & 94.8 & 92.8 & 95.3 & 97.7 & 96.7 & \textbf{95.5} \\
      \textit{\color{gray} $\text{LayoutLMv3-Base}^{*\dagger}$}~\cite{huang2022layoutlmv3} & \textit{\color{gray} 133M} & \textit{\color{gray}94.5} & \textit{\color{gray}90.6} & \textit{\color{gray}95.5} & \textit{\color{gray} 97.9} & \textit{\color{gray} 97.0} & \textit{\color{gray} 95.1}\\
      \textit{\color{gray} $\text{StructTextv2-Large}^{*\dagger}$} ~\cite{yu2023structextv2} & \textit{\color{gray} 238M} & \textit{\color{gray} -} & \textit{\color{gray} -}& \textit{\color{gray} -} & \textit{\color{gray} -} & \textit{\color{gray} -} & \textit{\color{gray} 95.5}\\
      
      \midrule
      \method & 85M & 94.4 & 89.5& 95.7 & 97.7& 97 & 94.9\\
      \bottomrule
      \end{tabular}}
      \caption{
      Document Layout Analysis mAP @ IOU [0.50:0.95] on PubLayNet validation set. Best overall result in \textbf{bold}. $^\dagger$StructTextv2 and LayoutLMv3 adopt longer \textbf{finetuning} schedules on PublayNet compared to the remaining baselines ($\approx 6\text{x}$ and $\approx 2\text{x}$ respectively). $^*$ Uses OCR as input during inference.}
      \label{tab:publaynet}
\end{table}

\subsection{Document Layout Analysis}
\label{sec:document-layout-analysis}
Document layout analysis (DLA) involves the detection of layouts of unstructured digital documents. This task helps identify elements such as \textit{tables, figures}, and other different types of textual layout elements like \textit{date, figure name, etc}. This task is crucial as it helps parse the documents for numerous downstream applications. We model DLA as an object detection problem, detecting elements of various classes with bounding boxes, using two popular document layout analysis datasets, PubLayNet~\cite{zhong2019publaynet} and D\textsuperscript{4}LA~\cite{da2023vgt} to evaluate performance on this task. 

\par
For object detection, we use a Cascade R-CNN~\cite{cai2019cascadercnn} as the detection pipeline on top of the backbone models, using the Detectron2~\cite{wu2019detectron2} library to evaluate our models. We use the same FPN and data processing setup as DiT~\cite{li2022dit} and VGT~\cite{da2023vgt}, with resolution-modifying modules at four different transformer blocks ($3$, $5$, $7$, and $11$) to adapt the single-scale ViT to the multi-scale FPN. Let $d$ be the total number of blocks; the $d/3^{rd}$ block is upsampled by $4\times$ using a module with $2$ stride-two $2\times2$ transposed convolution. For the output of the ${d/2}^{th}$ block, we use a single stride-two $2\times2$ transposed convolution to upsample by $2\times$. The output of the ${2d/3}^{th}$ block is utilized without additional operations. Finally, the output of $d^{th}$ block is downsampled by $2\times$ with stride-two $2\times2$ max pooling. All images are cropped with probability $0.5$ to a random rectangular patch which is then resized again such that the shortest side is at least $480$ and at most $800$ pixels while the longest is at most $1,333$ pixels. 

\subsubsection{PubLayNet}
PubLayNet~\cite{zhong2019publaynet} is a large dataset of 360K images for document layout analysis, created from over one million scientific articles in PubMed Central. It includes labeled images with five layout regions: \textit{text, title, list, figure, and table}. We finetune our model on the training split $(335,703)$ and evaluate on the validation split $(11,245)$. We follow the setting of DiT~\cite{li2022dit} and train for 60K steps with a batch size of $16$ and a learning rate of $4e-4$. 

\xhdr{Baselines and Results}
 We report the category-wise and overall mean average precision mAP@IoU$[0.50:0.95]$ of bounding boxes in Table~\ref{tab:publaynet}. We compare with vision-only input models such as DiT and Donut, as well as vision+OCR input models like VGT, LayoutLMv3.~\method~ achieves 94.9 which is on-par with DiT-L despite being less than ${1/3}^{rd}$ in model size and with less than ${1/2}$ of the pre-training steps. Our method also remains competitive with VGT despite not requiring OCR during inference, lower model size ($85$M as compared to $174$M in VGT). 
 From our observations, classes like \textit{text, list and table} are more ambiguous on textual semantics and might not be the best place to test our patch-text alignment loss. We see better performance gains on the D$^4$LA and M$^6$Doc benchmarks which has classes with more semantic distinction in the following sections.
 
\begin{table*}[h!]
    \begin{minipage}{0.45\textwidth}
    \setlength{\tabcolsep}{2pt}
    \scalebox{0.62}{
    \begin{tabular}{|l|c|c|c|c|c|c|c|}
    \toprule
    \textbf{Model}      & \textbf{DocTitle} & \textbf{ListText} & \textbf{LetterHead} & \textbf{Question} & \textbf{RegionList} & \textbf{TableName} & \textbf{FigureName}\\ 
    \midrule
    DiT-B~\cite{li2022dit} & 70.83 & 69.52 & 82.71 & 74.09 & 78.8 & 65.29 &  55.04\\
    DiT-L~\cite{li2022dit} & 72.13 & 68.73& 83.27& 75.1& 76.99& 65.93& 48.99\\
    VGT$^*$~\cite{da2023vgt} & 69.89 & 68.28 & \textbf{83.0} & 72.53 & \textbf{81.21} & 65.61 & 54.85 \\
    \method & \textbf{73.11} & \textbf{72.46} & 82.07 & \textbf{77.42} & 79.32 & \textbf{67.08} & \textbf{56.86} \\
    \midrule
    \textbf{Model}  &\textbf{Footer} & \textbf{Number}     & \textbf{ParaTitle} & \textbf{RegionTitle} & \textbf{LetterDear} & \textbf{OtherText} & \textbf{Abstract} \\ 
    \midrule
    DiT-B & 77.87 & \textbf{83.86} & 61.12 & 65.05 & 73.33 & 58.28 &  70.56\\
    DiT-L&76.76& 83.12& 60.9& 65.11& 72.88& 57.14&69.45\\
    VGT$^*$ & \textbf{79.0} & 82.71 & 61.11 & 64.39 & \textbf{75.08} & 57.97 & 74.9 \\

    \method & 77.88 & 83.15 & \textbf{64.07} & \textbf{65.17} & 72.7 & \textbf{61.25} & \textbf{78.25}\\
    \midrule
    \textbf{Model}   & \textbf{Table} & \textbf{Equation} & \textbf{PageHeader} & \textbf{Catalog}    & \textbf{ParaText} & \textbf{Date} & \textbf{LetterSign} \\ 
    \midrule
    DiT-B & 86.24 & 34.83 & 54.22 & 38.42 & 83.89 & 66.74 & 72.99\\
    DiT-L &87.18& 31.79& 55.1&49.08& 84.99&68.49&74.08\\
    VGT$^*$ & 86.4 & \textbf{49.0} & 52.28 & 49.37 & 84.89 & 67.88 & 74.01 \\

    \method & \textbf{86.9} & 32.26 & \textbf{58.22} & \textbf{60.98} & \textbf{85.75} & \textbf{71.4} & \textbf{76.31} \\
    \midrule
    \textbf{Model} & \textbf{RegionKV} & \textbf{Author} & \textbf{Figure} & \textbf{Reference} & \textbf{PageFooter} & \textbf{PageNumber} & \colorbox{gray!40}{\textbf{mAP}} \\
    \midrule
    DiT-B & 64.71 & 66.18 &75.64  & 81.46 & 65.78 & 58.60 & 68.0 \\
     DiT-L &67.07&66.04& 75.13&84.72&67.16&58.63&68.38\\
    VGT$^*$ & 66.56 & 64.09 & 76.65 & 84.19 & 64.14 &58.24  &  69.19\\
    \method & \textbf{70.3} & \textbf{70.66} & \textbf{75.73} & \textbf{84.45} & \textbf{65.82} & \textbf{60.64} & \textbf{70.72} \\
    \bottomrule
    \end{tabular}}
\caption{\small Performance comparison of different models across various document components of D$^4$LA benchmark. $^*$ Uses OCR as input during inference.}
\end{minipage}
 \hspace{0.08\textwidth}
\begin{minipage}{0.4\textwidth}
\centering
\label{tab:results}
\setlength{\tabcolsep}{3pt}
  \scalebox{0.65}{
  
  \begin{tabular}{lcccc}
  
    \toprule
    Model & \#Param & Precision & Recall & F1 \\
    \midrule
    Faster R-CNN & & 70.4 & 84.8& 76.0 \\
    ResNeXt-101d~\cite{xie2017aggregatedresidualtransformationsdeep}& & 93.87& 92.29 & 93.07 \\
    DeiT-B~\cite{touvron2021trainingdataefficientimagetransformers} & 87M & 94.29& 92.37 & 93.32 \\
    BEiT-B ~\cite{bao2021beit} & 87M & 94.12& 92.63& 93.37\\
    MAE-B~\cite{he2022mae} & 87M & 94.41 & 93.21 & 93.81\\
    DiT-B~\cite{li2022dit} & 87M & 94.70 & 93.07& 93.88\\
    DiT-L~\cite{li2022dit} & 304M & 94.52 & 93.36& 93.93 \\
    \midrule
    \method & 85M & \textbf{95.29} & \textbf{94.18} &  \textbf{94.73}\\
    \bottomrule
  \end{tabular}}
  \caption{ \small 
  Text detection accuracy (IoU@0.5) on FUNSD, where Mask R-CNN is used with different backbones. Best result in each category is indicated in \textbf{bold}}
    
\end{minipage}
\label{tab:d4la}
\end{table*}

\subsubsection{D\textsuperscript{4}LA}
\label{sec:d4la}
This dataset was introduced by VGT~\cite{da2023vgt}, containing around $12$K images with rich layouts that are manually annotated. It contains a lot more fine-grained classes than PubLayNet with a wider variation in the object sizes as well as objects that are distinguishable by the text present in them. The list of classes is available in Table~\ref{tab:results}. This makes it a more semantically challenging benchmark for document layout analysis. We use the same FPN and pre-processing setup as mentioned previously, and finetune all models for $60$K steps with a batch size of $12$ and learning rate of $2e-4$ with a warmup of $100$ steps. 

\xhdr{Baselines and Results} 
We report the category-wise and overall mean average precision mAP@IoU[$0.50:0.95$] of bounding boxes in Table~\ref{tab:results}. We compare against DiT and VGT, the current state-of-the-art baselines. Both baselines were fine-tuned with the same setup and hyperparameters as our method. ~\method~ sets a new SOTA (\textbf{$69.2\rightarrow70.72$}) on this benchmark, despite having less than half the parameters (85M vs 174M) and pre-training at a lower resolution (512 vs. 768) compared to VGT. In particular, \method~shows a marked improvement in object categories such as \emph{DocTitle, Question, ParaTitle, RegionTitle, RegionKV, Date, Author, and PageNumber}. These are highly fine-grained categories, where semantic understanding of the text is crucial, highlighting the efficacy of the proposed patch-text alignment loss. We do notice a tangible performance dip ($>1\%$) in classes such as \emph{Equation, RegionList, LetterDear and LetterHead}. These objects while holding individual semantic meanings could also be considered as subclasses of paragraph or list items, which might be a reason for their improper classification. We also observe a significant class imbalance for objects like \emph{equation} which results in a wide variation in performance. Deeper analysis on the predictions of~\method~ on \emph{equation} class revealed that ~\method~ was able to detect chemical equations which were originally not present in ground truth and was not predicted by VGT. We present and study qualitative examples of such cases in Appendix~\ref{app:failure}.

\begin{table}[!htp]
    \centering
    \scalebox{0.8}{
    \begin{tabular}{lcc|ccc}
    \toprule
           Model & Patch-Text  & Masking  & RVL-CDIP & PubLayNet & D\textsuperscript{4}LA  \\
          & Alignment & Ratio\\
     \midrule
          CLIP & -  & - & 90.97 & 93.3 & 64.5\\
          \method & - & 0.6 & 92.3& 94.35& 66.7\\ 
          \method & \checkmark & - & 92.51 & 94.35& 67.48 \\      
          \method & \checkmark & 0.6 & \textbf{92.84}& \textbf{94.62}& \textbf{67.92} \\ 
     \bottomrule
    \end{tabular}}
    \caption{Evaluation of performance with loss combinations. Evaluations are done at $224$ resolution at 160k pre-training steps. Publaynet and D\textsuperscript{4}LA were evaluated with default DiT config. Best results are highlighted in \textbf{bold}.}
    \label{tab:losses}
\end{table}
\begin{table}[h!]
    \centering
    \scalebox{0.8}{
    \begin{tabular}{c|ccc}
    \toprule
         Masking Ratio & RVL-CDIP & PubLayNet & D\textsuperscript{4}LA  \\
     \midrule
          0.2 & 92.54& 94.43& 67.74\\ 
          0.4 & 92.79& 94.54& 67.7\\        
         0.6 & \textbf{92.84}& \textbf{94.62}& \textbf{67.92} \\ 
     \bottomrule
    \end{tabular}}
    \caption{Evaluation of performance at different masking ratios at 160k pre-training steps. Best results are highlighted in \textbf{bold}.}
    \label{tab:masking_ratios}
\end{table}
\begin{table}[h!]
    \centering
    \setlength{\tabcolsep}{3pt}
    \scalebox{0.8}{
    \begin{tabular}{l  c c c}
        \toprule
        \textbf{Method}  & \textbf{AP50} & \textbf{AP75} & \textbf{mAP} \\
        \midrule
        Mask R-CNN   & 58.4 & 46.2  & 40.1 \\
        Cascade R-CNN   & 70.5 & 62.9  & 54.4 \\
        HTC  & 74.3 & 67.2  & 58.2 \\
        SCNet   & 73.5 & 65.1  & 56.1 \\
        Deformable DETR   & 76.8 & 63.4 & 57.2 \\
        ISTR   & 80.8 & 70.8  & 62.7 \\
        TransDLANet & 82.7 & 72.7 & 64.5 \\
        VSR   & 76.2 & 68.8  & 59.9 \\
        DINO   & 84.6 & 76.7  & 68.0 \\
        M2Doc$^{*}$  & 78.0 & 70.7 & 61.8 \\
        DiT-B  & 84 & 76.4  & 67.6\\
        \midrule
        DoPTA  & \textbf{85.8} & \textbf{78.5} & \textbf{69.5} \\
        \bottomrule
    \end{tabular}}
    \caption{Performance comparison of different methods on M$^6$Doc benchmark.$^*$ Uses OCR as input during inference.Best result in each category is indicated in \textbf{bold}}
    \label{tab:results}
\end{table}

\subsubsection{M\textsuperscript{6}Doc}
\label{sec:m6doc}
M$^6$Doc is another layout detection benchmark with highly nuanced object classes like \textit{poem}, \textit{examinee information}, \textit{weather forecast} that could rely on semantic understanding for efficient detection. The dataset constitutes of $9080$ images with $76$ highly diverse object classes. There are also objects like \textit{QR Code}, \textit{flag}, \textit{underscore} which are more visually distinguishable. We finetune our model for $90$K steps with a batch size of 16 and a learning rate of $4e-4$ with a warmup of $100$ steps using the Cascade-RCNN framework.\\
\xhdr{Baselines and Results} We report the AP50, AP75 and mAP scores of our model and baselines on the validation set. Except DiT-B, all the baselines are reported from M2Doc\cite{zhang2024m2doc}. We train DiT-B with the same hyperparameters as ours. ~\method~beats DiT-B by \textbf{+1.9 mAP}~\method~achieves \textbf{$69.5$} mAP setting SOTA on the benchmark. 

\subsection{Text Detection} 
\label{sec:text-detection}
We test the word-level text detection capability of~\method~encoder using the FUNSD~\cite{jaume2019funsddatasetformunderstanding} dataset. It is a subset of the RVLCDIP dataset constructed to perform form understanding tasks like text detection, entity labeling, and information extraction. The dataset comprises of 199 annotated images (149 train and 50 test images). Following DiT~\cite{li2022dit}, we employ mask R-CNN framework to perform the text detection using~\method~encoder
as the backbone. We vary the anchor box sizes from the previous experiments to $[4,8,16,32,64]$ as the expected predictions are smaller in size compared to paragraph-level predictions earlier. The learning rate is set to $1e-4$ with a batch size of 16 and finetuning is performed for 60k steps, following DiT. This setup is followed for all baselines. The resolution and data augmentation is the same as the document layout analysis setup, described in Sec.~\ref{sec:document-layout-analysis}. 

\xhdr{Baselines and Results} We compare against various CNN and ViT backbones and report the precision, recall, and F1 Score at IoU=$0.5$. We do not compare against VGT since it uses the OCR as an input, making the task redundant. StructTextv2 is omitted due to non-availability of code/weights.~\method~outperforms the previous best result from DiT-L setting new SOTA, despite pre-training for only $1/2$ the pre-training steps, and having less than $1/3^{rd}$ the parameters. 

\subsection{Inference Time Analysis}

In this section, we analyze the performance benefits of our OCR-free inference setting compared to baselines on layout detection models on the D4LA test set. We observe that OCR parsing with EasyOCR\cite{JadedAI} takes an average of 1.02 seconds per image. Since VGT relies on OCR during inference, it operates at 0.81 FPS while our ~\method~ achieves 9.56 FPS which is —\textbf{12×} faster than VGT. Not only does ~\method~ achieve SOTA performance, but it also has significantly improved inference speed. A visual comparison is provided in Figure~\ref{fig:fps}. All the methods were carefully tested on the same A100 gpus.

\section{Ablation Study}
\label{sec:ablations}

In this section, we study the effect of the individual components of~\method. It is crucial to study the performance of our proposed patch-text contrastive loss. The reconstruction loss is also an important component, bringing visual information to the features where textual features are not available. To understand the contribution of each loss objective independently, we evaluate our method on an array of different masking ratios, as well as in the absence of the reconstruction loss and patch-text contrastive loss. We also compare with the CLIP model, to quantify the contribution of the architecture, in the absence of our loss components. All experiments in the ablation study were carried out while pre-training for 10 epochs on the IIT-CDIP Dataset. The batch size, learning rate, data augmentation, and other hyperparameters were kept the same as the original setup. We evaluate the performance by benchmarking on document image classification and document layout analysis. The setup and fine-tuning parameters for each downstream evaluation are unchanged from Sec.~\ref{sec:experiments}.

\xhdr{Results}
The results of individual loss components are summarized in Table~\ref{tab:losses} and the effect of masking ratio in Table ~\ref{tab:masking_ratios}. It is clear that both loss components signifcantly improve performance over the Row 1 baseline. Further, as seen in Row 3, the model trained only with the proposed patch-text contrastive loss retains strong performance. In particular, this variant outperforms reconstruction only training on RVL-CDIP and D$4$LA, while matching performance on PubLayNet. This result highlights the efficacy of the patch-text contrastive loss in learning effective visual representations for document layout analysis. The results see a clear improvement when both the losses are included. In Table~\ref{tab:masking_ratios} we see an improvement of $0.3-0.6$ performance points in all benchmark scores as the masking ratio is increased. The combination of patch-text alignment and reconstruction objectives enables the model to learn strong visual representations that generalize across various task settings. 


\section{Conclusion and Future Work}
\label{sec:conclusion}

In this work, we extend fine-grained image-text alignment to document images via a novel patch-text alignment objective. Our work shows the efficacy of leveraging the textual information in document images to solve visual tasks, which is still an underexplored direction. We combine this novel objective with a masked reconstruction loss to build \method, a strong document encoder model that achieves state-of-the-art results across document image classification, layout analysis, and text detection tasks, consistently outperforming baselines that use larger models, extra information (OCR) as input, and longer pre-training schedules. We hope that this work motivates further research into methods that can leverage text in document images for \textit{visual understanding}. 

Our work opens several new avenues for further exploration. 
We aim at extending  ~\method~to newer architectures such as SwinTransformer, which could provide better results for document images with small objects and details, exploring alternative strategies for text masking, and leveraging synthetic data generation techniques to increase the size and diversity of the training dataset. We aim to explore these directions as part of our future work.


{
    \small
    \bibliographystyle{ieeenat_fullname}
    \bibliography{main}
}

\clearpage
\setcounter{page}{1}
\maketitlesupplementary
\appendix
The appendix is structured as follows - In Appendix~\ref{app:pixparse} we present and discuss results of pre-training~\method~on the PixParse dataset. Appendix~\ref{app:examples} contains additional qualitative examples of the effect of our various pre-training objectives. In Appendix~\ref{app:failure} we qualitatively analyse the relative lower performance of~\method~on certain categories in D\textsuperscript{4}LA. 

\section{Results on PixParse}
\label{app:pixparse}
While the pre-training of~\method~ was carried out using the CDIP dataset due to its scale, we also explore the use of the PixParse-PDFA\footnote{\href{https://huggingface.co/datasets/pixparse/pdfa-eng-wds}{https://huggingface.co/datasets/pixparse/pdfa-eng-wds}} dataset for pretraining~\method~in this section. The born-digital nature of this dataset, ensuring high quality OCR information without any OCR errors makes it a high-quality source of pre-training data for~\method. However, this dataset is much smaller (19M pages) than the CDIP dataset, and filtering to remove documents with bad aspect ratios further reduces its number, preventing its use as the primary dataset. Due to this reason, and to maintain parity with the baselines, we chose the CDIP dataset for pretraining~\method. 

In Table~\ref{tab:pixparse}, we report the results of pre-training~\method~on the PixParse dataset, and compare it to the variant trained on CDIP. All hyperparameters for~\method~ are identical to the original pre-training setting outlined in Sec.~\ref{sec:experiments}. However, we only train for $80$K steps as we find this sufficient to notice significant differences between pre-training on PixParse and CDIP. We notice a consistent trend, wherein~\method~pre-trained on PixParse achieves consistently lower scores than the CDIP variant across all benchmarks. Despite the high quality OCR data, this may be caused due to to major reasons - i) The low number of samples in PixParse, leading to a larger degree of overfitting on the pre-training dataset, and ii) The distribution of downstream benchmarks like RVL-CDIP and FUNSD more closely matching that of the pre-training data in CDIP, as these are both datasets comprising of scanned documents, which may prove to be slightly OOD when pre-training on PixParse.

\begin{table}[h!]
    \setlength{\tabcolsep}{4pt}
    \centering
    \begin{tabular}{cc|ccc}
    \toprule
         Model & Dataset & RVL-CDIP & D\textsuperscript{4}LA & FUNSD \\
    \midrule
         ~\method & CDIP & \textbf{93.78} & \textbf{69.69} & \textbf{94.31} \\ 
         \rowcolor{gray!30}
         ~\method & PixParse & 93.47 & 68.77 & 94.19 \\ 
    \bottomrule
    \end{tabular}
    \caption{\method~trained on different pre-training datasets for $80$K steps. Across all setting and benchmarks,~\method~pre-trained on CDIP outperforms the PixParse variant consistently. Best result in each category (regular and high-resolution) in \textbf{bold}.}
    \label{tab:pixparse}
\end{table}

\section{Additional Qualitative Examples}
\label{app:examples}
\begin{figure*}
    \centering
    \begin{subfigure}[b]{0.45\linewidth}
        \centering
        \includegraphics[width=\linewidth]{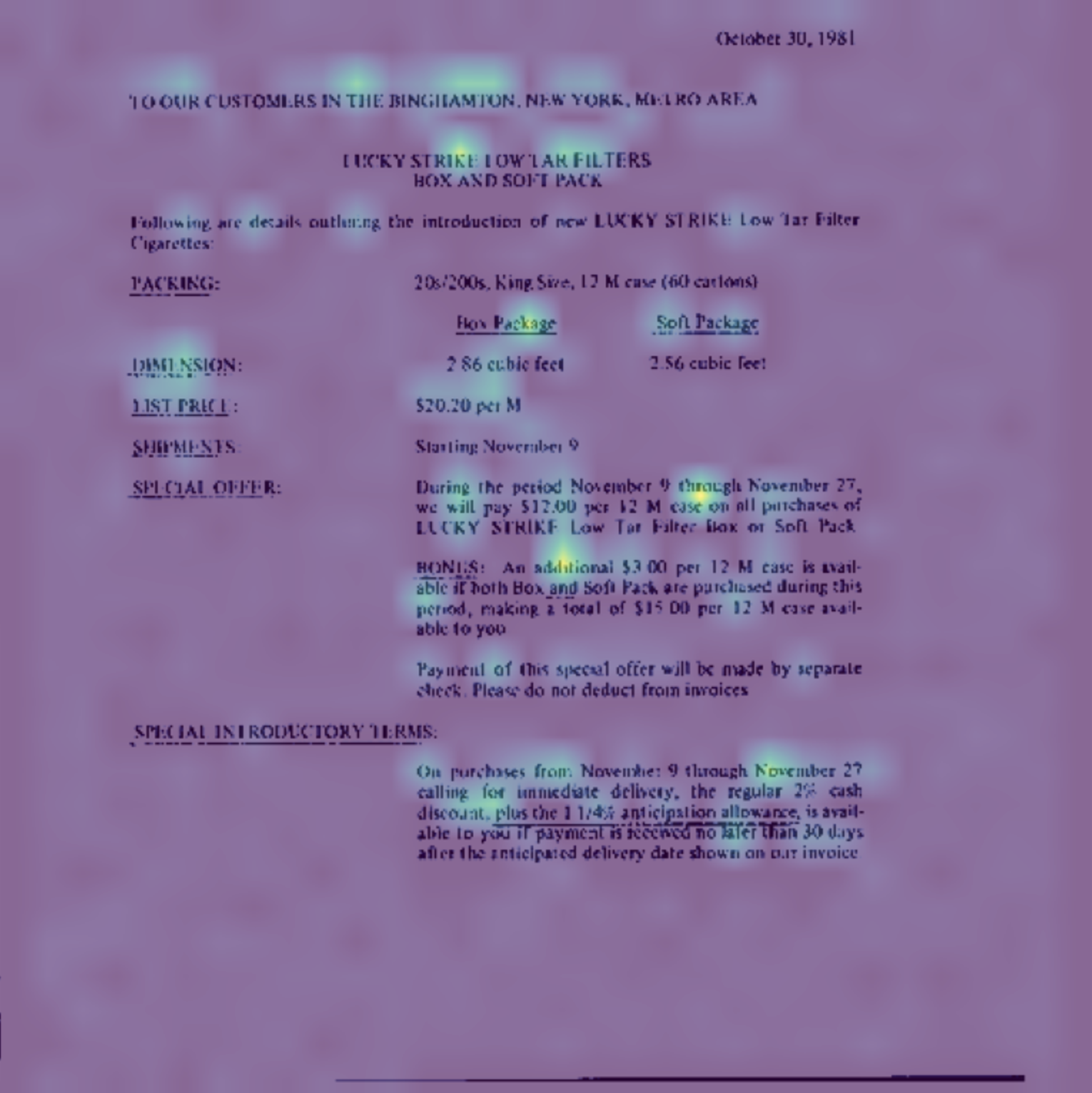}
        \caption{`Additional'}
    \end{subfigure}
    \hfill
    \begin{subfigure}[b]{0.45\linewidth}
        \centering
        \includegraphics[width=\linewidth]{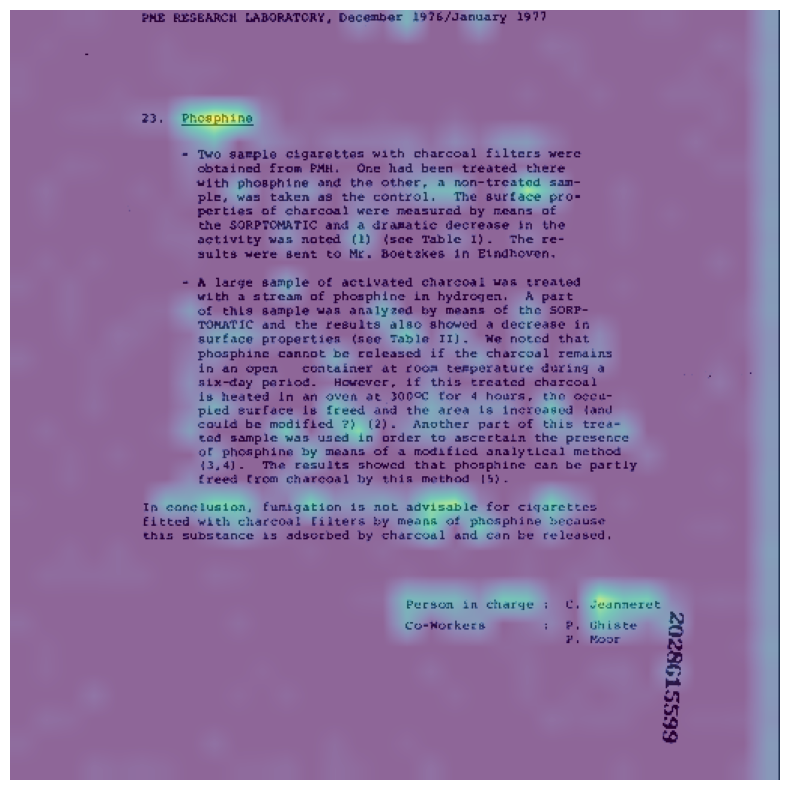}
        \caption{`Phosphine'}
    \end{subfigure}
    \vskip\baselineskip
    \begin{subfigure}[b]{0.45\linewidth}
        \centering
        \includegraphics[width=\linewidth]{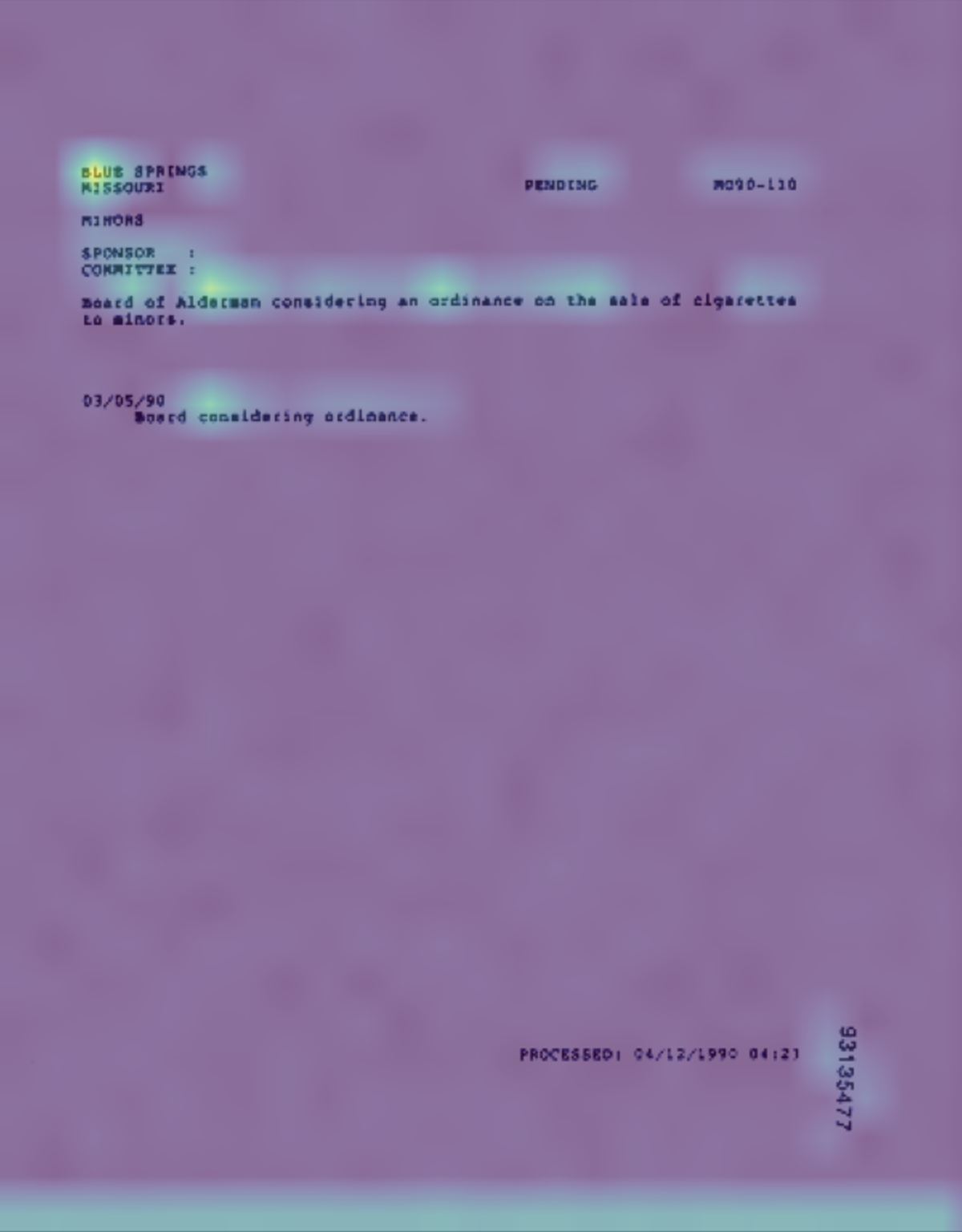}
        \caption{`Blue'}
    \end{subfigure}
    \hfill
    \begin{subfigure}[b]{0.45\linewidth}
        \centering
        \includegraphics[width=\linewidth]{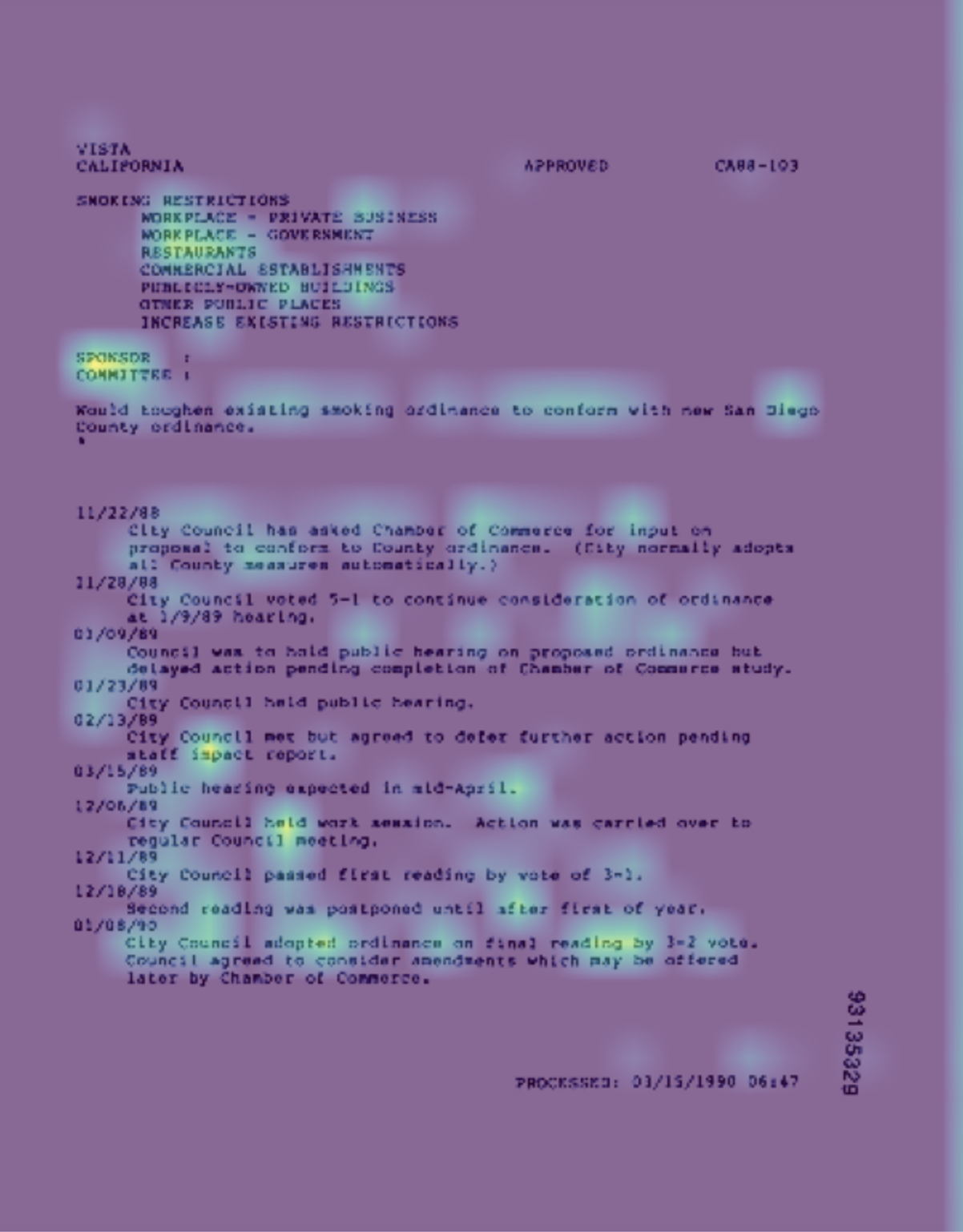}
        \caption{`Sponsor'}
    \end{subfigure}
    \caption{Heatmap visualisation of the normalised dot product similarity of image patch embeddings and text embeddings taken from the~\method~model, demonstrating its ability to find individual words in document images. Despite some noise, there is a clear spike in dot product similarity at the appropriate text region. The target text word for each image is mentioned.}
    \label{fig:additional_attention}
\end{figure*}

In Fig.~\ref{fig:additional_attention}, we demonstrate additional qualitative examples of the effect of the patch-text alignment objective on the~\method~encoder. The results demonstrate the ability of the pretrained~\method~ encoder to isolate individual words in a document image, which was the goal of the patch-text alignment objective. This ability translates to better performance on downstream benchmarks, as demonstrated in Sec.~\ref{sec:experiments}.

\section{Analysis of Failure Cases}
\label{app:failure}
\begin{figure*}
    \centering
    \includegraphics[width=\linewidth]{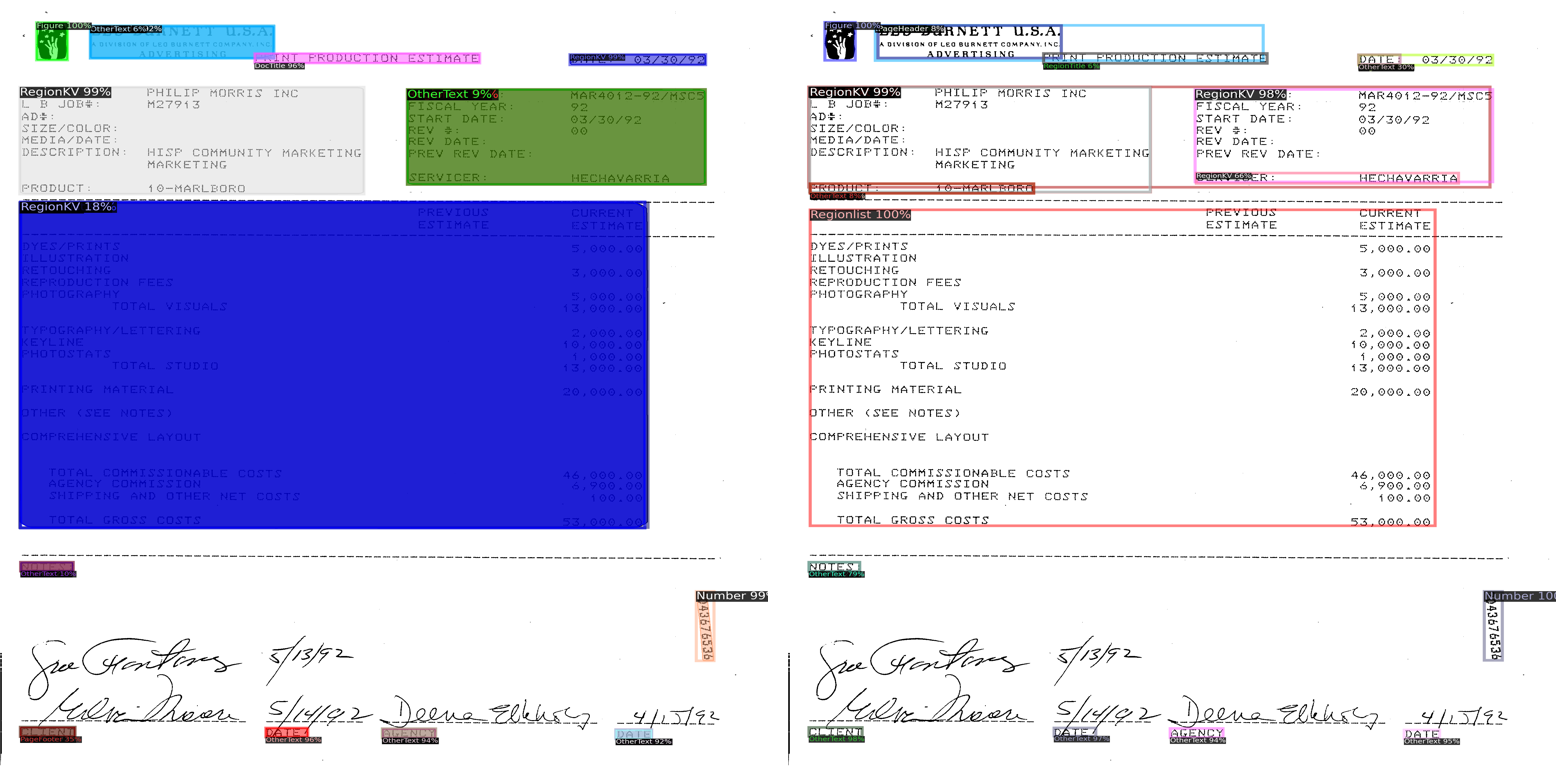}
    \caption{Failure case of~\method on layout analysis on D\textsuperscript{4}LA benchmark. Left is~\method. Right is VGT.~\method incorrectly marks the central region as \textit{RegionKV}, which was found to be a common error mode.}
    \label{fig:failure_regionlist}
\end{figure*}

\begin{figure*}
    \centering
    \includegraphics[width=\linewidth]{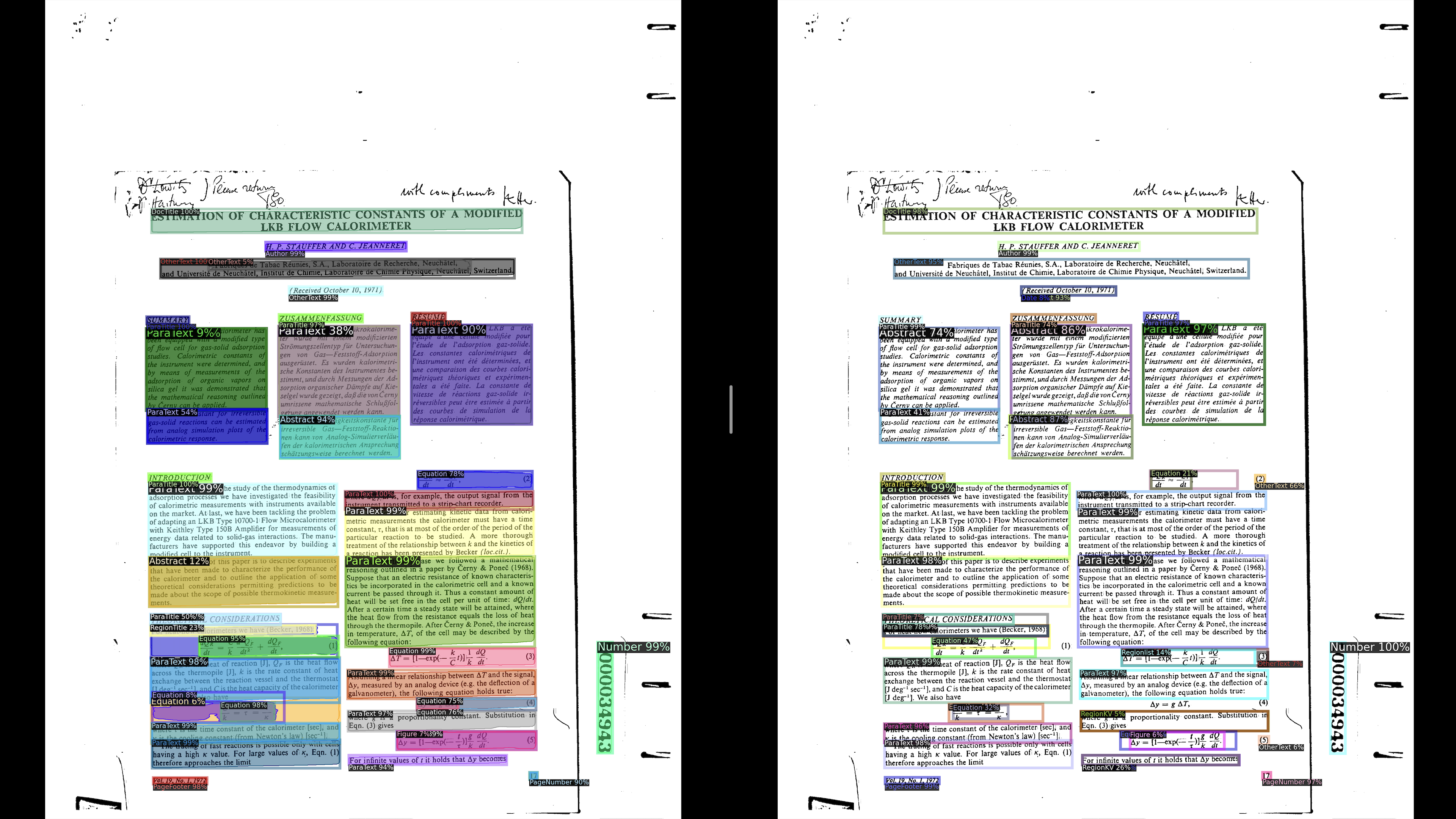}
    \caption{Prediction from ~\method on layout analysis on \textit{Equation} class. Left is~\method. Right is VGT.~\method identifies equation objects that were not identified by VGT but also encloses extraneous regions leading to poor performance.}
    \label{fig:failure_equation}
\end{figure*}

\begin{figure*}
    \centering
    \includegraphics[width=\linewidth]{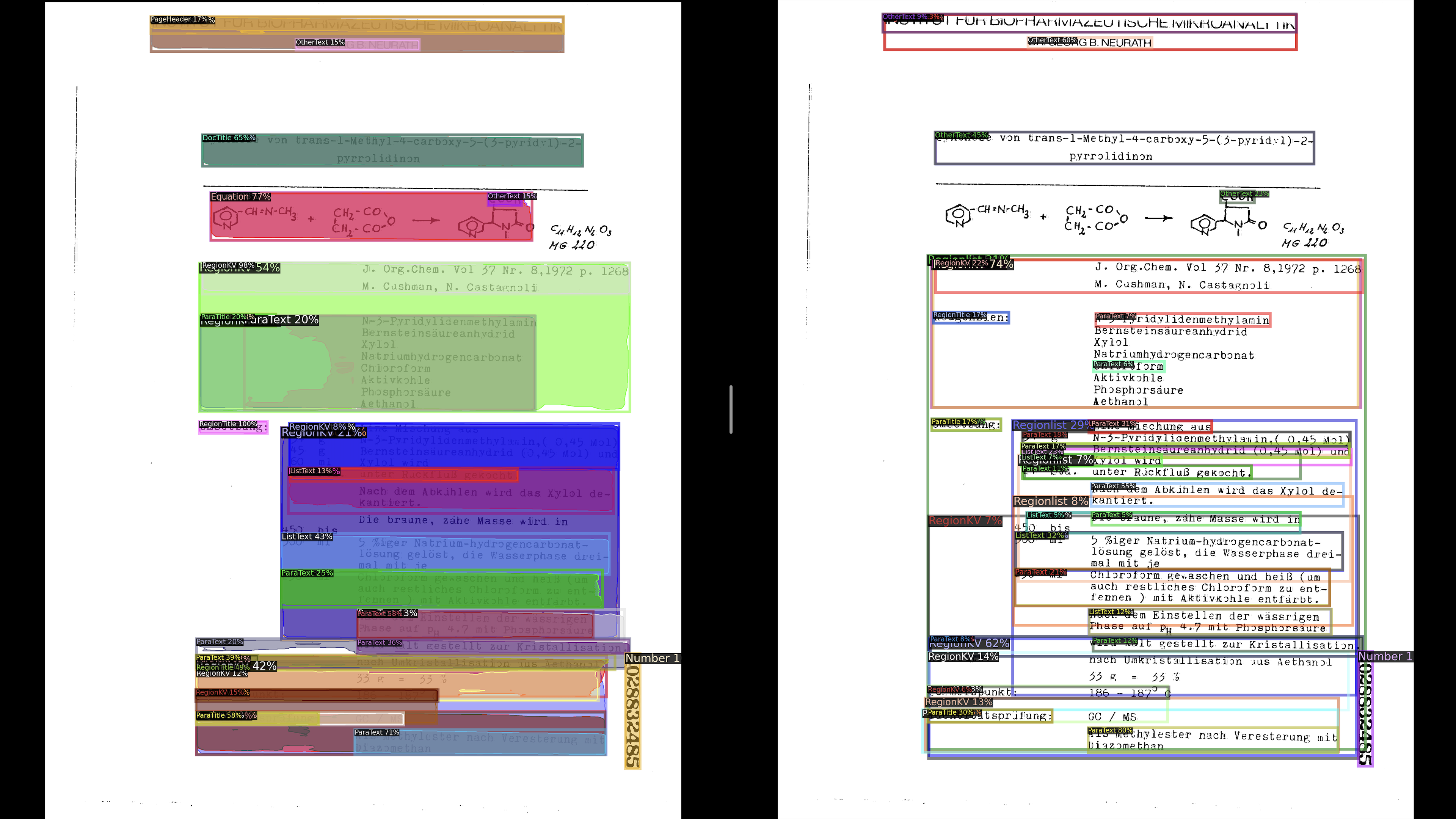}
    \caption{Prediction from ~\method on layout analysis on \textit{Equation} class. Left is~\method. Right is VGT.~\method identifies a chemical equation objects that were not identified by VGT.}
    \label{fig:failure_equation2}
\end{figure*}

In this section, we analyze the lower performance of~\method~on certain classes in the D\textsuperscript{4}LA dataset. In particular, we observe lower performance on the \textit{RegionList} category. We found that this occurs due to a common error made by~\method~, as demonstrated in Fig.~\ref{fig:failure_regionlist}, where the model incorrectly marks \textit{RegionList} as \textit{RegionKV}. This is most likely due to the high visual similarity between the two classes, and the ground truth labels often seem to be ambiguous. Another area of low performance was the \textit{Equation} category, where~\method~ ($32.26$ mAP) yields far lower performance than VGT ($49.0$ mAP). We identify that this category has an extremely low occurence in the dataset (only 2/3 samples in total), which may explain the low performance. We did not observe any other consistent trend which may explain said low performance. ~\method~ however demonstrates interesting performance on \textit{equation} class as illustrated in Fig.\ref{fig:failure_equation2} and \ref{fig:failure_equation} where we observe ~\method~predicting equation entities missed by VGT but with incorrect boundaries leading to poor performance. ~\method~also predicts chemical equation which was missed by the VGT. We identified that the very high class imbalance in the total dataset led to high variation in the final mAP performance.

\begin{figure}
    \centering
    \scalebox{0.9}{\includegraphics[width=7cm]{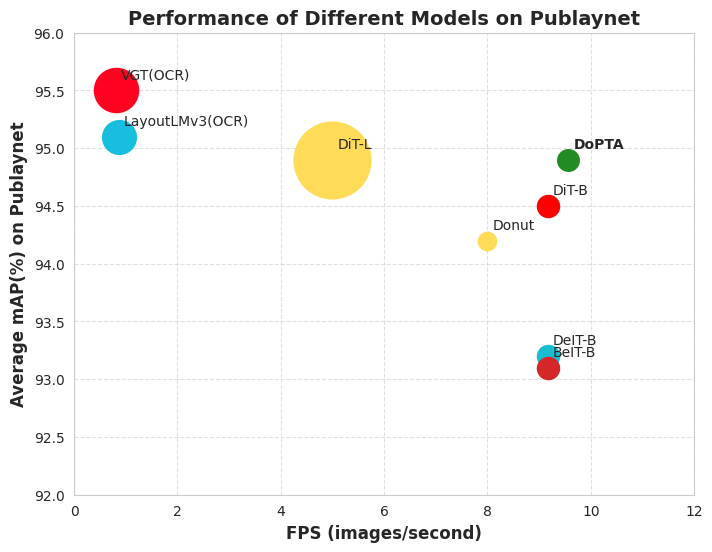}}
    \caption{Plot explaining FPS and Publaynet accuracy of various models. Model(OCR) denotes the FPS when OCR parsing is taken into account for computing inference time.}
    \label{fig:fps}
\end{figure}


\end{document}